\documentclass[sigconf,authorversion]{acmart}

\AtBeginDocument{%
  \providecommand\BibTeX{{%
    \normalfont B\kern-0.5em{\scshape i\kern-0.25em b}\kern-0.8em\TeX}}}

\setcopyright{acmcopyright}
\copyrightyear{2021}

\acmConference[KDD'21]{KDD'21: 27th ACM SIGKDD Conference on Knowledge Discovery and Data Mining}{August 14--18, 2021}{Virtual Event, USA}

\copyrightyear{2021}
\acmYear{2021}
\setcopyright{acmlicensed}\acmConference[KDD '21]{Proceedings of the 27th ACM SIGKDD Conference on Knowledge Discovery and Data Mining}{August 14--18, 2021}{Virtual Event, Singapore}
\acmBooktitle{Proceedings of the 27th ACM SIGKDD Conference on Knowledge Discovery and Data Mining (KDD '21), August 14--18, 2021, Virtual Event, Singapore}
\acmPrice{15.00}
\acmDOI{10.1145/3447548.3467166}
\acmISBN{978-1-4503-8332-5/21/08}

\usepackage{graphicx}
\graphicspath{ {./extended/} }
\usepackage{algorithm}
\usepackage[noend]{algpseudocode}
\usepackage{algorithmicx}

\usepackage{siunitx}
\usepackage[utf8]{inputenc} 
\usepackage[T1]{fontenc}    
\usepackage{hyperref}       
\usepackage{url}            
\usepackage{booktabs}       
\usepackage{amsfonts}       
\usepackage{nicefrac}       
\usepackage{microtype}      

\usepackage{textcomp}
\usepackage{mathtools}
\usepackage{mathrsfs}
\usepackage{multicol}
\usepackage{caption}
\usepackage{subcaption}
\usepackage{chngcntr}

\usepackage{balance}

\newcommand{\KwInput}{\hspace*{\algorithmicindent}\textbf{Input:}}
\begin{document}

\title[]{TimeSHAP: Explaining Recurrent Models\\through Sequence Perturbations}

\author{Jo\~ao Bento}
\affiliation{%
  \institution{Feedzai}
  \institution{Instituto Superior T\'ecnico, ULisboa}
   \country{}}

\email{joao.bento@feedzai.com}


\author{Pedro Saleiro}
\affiliation{%
  \institution{Feedzai}
     \country{}}

\email{pedro.saleiro@feedzai.com}

\author{ Andr\'e F. Cruz}
\affiliation{%
  \institution{Feedzai}
     \country{}}
\email{andre.cruz@feedzai.com}

\author{Mário A.T. Figueiredo}
\affiliation{%
  \institution{Instituto Superior T\'ecnico, ULisboa}
   \institution{Instituto de Telecomunica\c{c}\~oes}
  \country{}}
\email{mario.figueiredo@tecnico.ulisboa.pt}

\author{Pedro Bizarro}
\affiliation{%
  \institution{Feedzai}
   \country{}}

\email{pedro.bizarro@feedzai.com}

\renewcommand{\shortauthors}{}

\begin{abstract}
Although recurrent neural networks (RNNs) are state-of-the-art in numerous sequential decision-making tasks, there has been little research on explaining their predictions.
In this work, we present TimeSHAP, a model-agnostic recurrent explainer that builds upon KernelSHAP and extends it to the sequential domain. TimeSHAP computes feature-, timestep-, and cell-level attributions. As sequences may be arbitrarily long, we further propose a pruning method that is shown to dramatically decrease both its computational cost and the variance of its attributions.
We use TimeSHAP to explain the predictions of a real-world bank account takeover fraud detection RNN model, and draw key insights from its explanations:
i) the model identifies important features and events aligned with what fraud analysts consider cues for account takeover;
ii) positive predicted sequences can be pruned to only 10\% of the original length, as older events have residual attribution values;
iii) the most recent input event of positive predictions only contributes on average to $41\%$ of the model's score;
iv) notably high attribution to client's age, suggesting a potential discriminatory reasoning, later confirmed as  higher false positive rates for older clients.

\end{abstract}

\keywords{TimeSHAP, RNN, SHAP, Shapley Values, XAI, Explainability}

\maketitle

\section{Introduction}
\label{sec:introduction}

\begin{figure}[b]
	\centering
    \includegraphics[width=0.95\columnwidth]{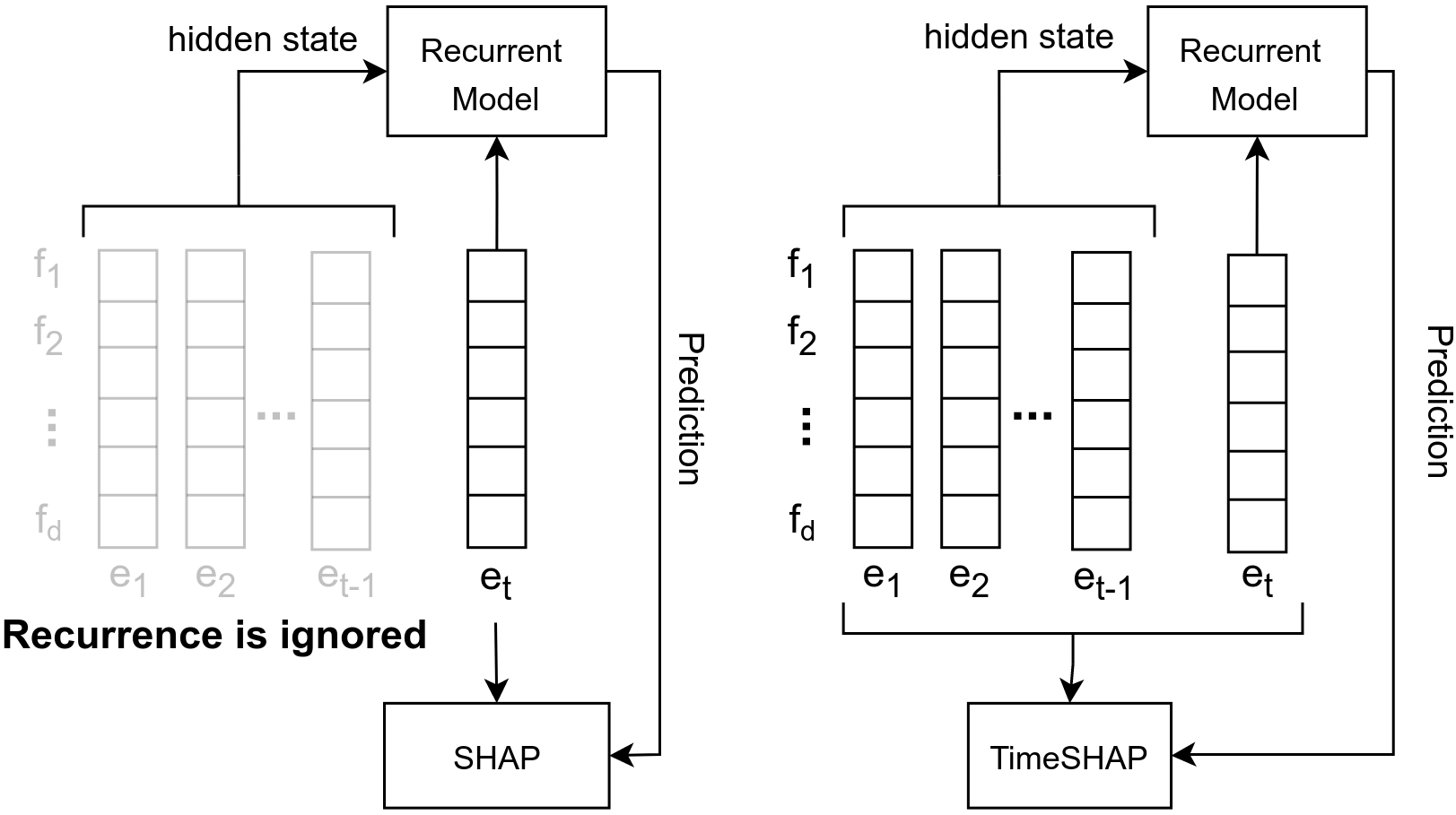}
	\caption{
	Current SHAP-based methods (left) only calculate attributions for a single input vector. TimeSHAP (right) applies perturbations throughout the input sequence.}
	
	\label{fig:chap1_shap}
\end{figure}

In recent years, numerous methods have been put forth to explain machine learning (ML) models~\cite{Simonyan2014DeepIC,denil2014extraction,li2016understanding,MONTAVON2017211,10.1371/journal.pone.0130140,shrikumar2019learning,lstmvis,murdoch2017automatic,karpathy2015visualizing,SHAPNIPS2017_7062}.
However, recurrent neural network (RNN)~\cite{rumelhart1986learning} models pose a distinct challenge, as their predictions are not only a function of the immediate input instance, but also of the previous input instances in the sequence and the context drawn thereafter (hidden state).  Blindly applying state-of-the-art explainers
to RNNs disregards the importance of past events and the features throughout the sequence, only attributing importance to features of the current input (as illustrated in Figure~\ref{fig:chap1_shap}).

In this work, we aim to explain RNN-based models applied to fraud detection. Identity fraud cost banks, merchants, and consumers a total of 16.9 billion USD in 2019~\cite{teddler2020javelin}, and it is expected to rise as criminals become more active during times of economic hardships. To prevent fraudulent transactions, banks and merchants are increasingly investing in ML-based fraud detection systems. To detect fraudulent behavior, ML models take as input the historic sequence of events of the entity of interest, such as all the past payments of a consumer credit card or the transaction history of a bank account. RNN-based models, such as long-short term memory (LSTM)~\cite{Hochreiter1997} and gated recurrent unit (GRU)~\cite{Cho2014}, are state-of-the-art models in these sequential tasks. However, the complex decision-making processes in these models is still seen as a black-box, creating a tension between accuracy and interpretability.

%
Understanding the decision-process of these black-boxes is paramount in high-stakes decisions that impact people’s lives,  such as identity fraud detection, otherwise, financial institutions and regulators lack trust, and refrain from deploying them to production. Due to business constraints, an explainer in this context must not be model-specific, as we need to support various deployment scenarios, including multiple fraud detection use cases, clients, and geographies. Moreover, we may only have access to the model through an inference API, restricting our approach to be perturbation-based, post-hoc, and model-agnostic.

Recently, Lundberg and Lee~\cite{SHAPNIPS2017_7062} unified a large number of explanation methods into a single family of ``additive feature attribution methods'', proposing also a model-agnostic, post-hoc explainer, dubbed KernelSHAP, which satisfies three desirable properties of explanations: \textit{consistency}, \textit{local accuracy}, and \textit{missingness}. Although KernelSHAP has become quite popular, it is not directly applicable to sequential decision-making tasks, as it only calculates attributions for features of a single input instance. 



To address this limitation, we propose TimeSHAP, a post-hoc model-agnostic recurrent explainer. TimeSHAP leverages the strong theoretical foundations and empirical results of KernelSHAP~\cite{SHAPNIPS2017_7062}, and adapts it to the recurrent setting. By doing so, we enable explaining, not only which features are most important to a recurrent model throughout a sequence, but also which previous events had the largest contribution to the the current prediction. 
As sequences are arbitrarily long, we further propose a temporal coalition pruning algorithm to dramatically reduce the computational cost of TimeSHAP, by aggregating older events, unimportant to the current prediction. Doing so, significantly improves the stability of TimeSHAP attributions and reduces the computational cost needed to obtain them.

We validate our method at Feedzai, on a real-world bank account takeover fraud detection model comprising a GRU layer and a feed-foward classifier (Section~\ref{sec:case_study}). Fraud analysts corroborate, through TimeSHAP explanations, the ability of the model to detect a typical account takeover pattern consisting of \textit{enrollment}-\textit{login}-\textit{transaction}. The temporal coalition pruning results reveal that account takeover fraud can be predicted based on a very short sub-sequence of the most recent events, with the last one only contributing, on average, to 41\% of the model score. Additionally, a notably high attribution to client's age suggested a potential disparity between different age groups, which was later confirmed on a bias audit.

In summary, the main contributions of this work are:
\begin{itemize}
    \item Adaptation of KernelSHAP~\cite{SHAPNIPS2017_7062} to sequences (Section~\ref{sec:Method});
    \item Calculation of three types of explanations: feature-, event-, and cell-wise explanations (Sections~\ref{sec:TimeSHAP} and ~\ref{subsec:cell_level});
    \item A pruning method that reduces the execution cost of TimeSHAP and decreases explanation variance (Section~\ref{subsec:pruning});
    \item Validation of our method on a real-world account takeover fraud detection use case (Section~\ref{sec:case_study}).
\end{itemize}


\section{Related Work}
\label{sec:RW}

Research on ML explainers can generally be subdivided into two categories: model-agnostic and model-specific.

\textbf{Model specific explainers} exploit characteristics of the model's inner workings or architecture to obtain more accurate explanations of its reasoning~\cite{Ribeiro2016}.
The task of explaining RNNs is often tackled by using attention mechanisms~\cite{NIPS2019_9264,Zhang_2018,RETAIN,attentionRNNs}.
However, whether attention can in fact explain a model's behavior is debatable and a known source of controversy in the ML community~\cite{serrano-smith-2019-attention,attention-is-not-explanation,attention-is-not-not-explanation}.

Deep learning (DL) models, in which RNNs are included, can also be explained using gradient-based methods.
These explainers attribute a weight $w_i$ to each feature, representing the importance, or saliency, of the i-th feature, based on the partial derivatives of the prediction function $f(x)$ with respect to the input $x_i$: $w_i = \left|\frac{\partial f(x)}{\partial x_i}\right|$ \cite{Simonyan2014DeepIC,denil2014extraction,axiomatic}.

Another approach in the class of gradient-based methods redefines the relevance-backpropagation rules instead of using the out-of-the-box gradient of the explained models. These methods iteratively backpropagate a relevance value (which initially corresponds to the predicted score) over all the layers of the model until the input layer is reached and the relevance is distributed across all features. The contribution that reaches each input feature is that feature's final attribution.
The rules to distribute the relevance at each layer depend on the domain in which the model is applied~\cite{MONTAVON2017211,10.1371/journal.pone.0130140,shrikumar2019learning}.
However, 
when explaining sequential inputs, DL-specific methods focus on each input, leaving event relevance as a largely unexplored research direction. Finally, this approach is susceptible to noise and complexities that may affect the gradient, such as the vanishing gradients problem~\cite{hochreiter1991untersuchungen,NIPS2019_9264}.
%

\textbf{Model-agnostic explainers} are substantially more flexible as they can explain any architecture and can be applied to already trained architectures, possibly in production.
These explainers generally rely on post-hoc access to a model's predictions under various settings, such as perturbations of its input~\cite{bbexplainers}.
A perturbation $h_x$ of the input vector $x \in \mathbb{X}^m$ is the result of converting all values of a coalition of features $z \in \left\{0, 1\right\}^m$ to the original input space $\mathbb{X}^m$, such that $z_i=1$ means that a feature $i$ takes its original value $x_i$, and $z_i=0$ means that a feature $i$ takes some uninformative background value $b_i$ representing its removal.
Hence, the input perturbation function $h_x$ is given by:

\begin{equation}
\label{eq:perturbation}
h_x(z) = x \odot z + b \odot (\mathbf{1} - z),
\end{equation}
where $\odot$ is the component-wise product. The vector $b \in \mathbb{X}^m$ represents an uninformative input sample, which is often taken to be the zero vector~\cite{LIME}, $b = \mathbf{0}$, or to be composed of the average feature values in the input dataset~\cite{SHAPNIPS2017_7062}, $b_i=\overline{x_i}$.

Lundberg and Lee~\cite{SHAPNIPS2017_7062} unified this and other explainers (both model-agnostic and model-specific) into a single family of ``additive feature attribution methods''.
Moreover, the authors prove that there is a single solution to this family of methods that fulfills both local accuracy (the explanation model should match the complex model locally), missingness (features that are set to be missing should have no impact on the predictions), and consistency (if a feature's contribution increases, then its attributed importance should not decrease).

Those authors put forth KernelSHAP~\cite{SHAPNIPS2017_7062}, a model-agnostic explainer that fulfills these three properties.
KernelSHAP approximates the local behavior of a complex model $f$ with a linear model of feature importance $g$, such that $g(z) \approx f(h_x(z))$.
The task of learning the explanation model $g$ is cast as a cooperative game where a reward ($f(x)$, the score of the original model) must be distributed fairly among the players ($i \in \left\{1, \dots, m\right\}$, the features).
The optimal reward distribution is given by the Shapley values formulation~\cite{Young1985}.
However, obtaining the exact Shapley values for all features would imply generating all possible coalitions of the input, $z \in \left\{0, 1\right\}^m$, which scales exponentially with $m$, the number of features in the model.
As this task is computationally intractable, KernelSHAP approximates the exact values by randomly sampling feature coalitions~\cite{Strumbelj2014}.
The authors further show that a single coalition weighing kernel, $\pi_x(z)$, and a single loss metric, $L(f, g, \pi_x)$, lead to optimal approximations of the Shapley values,

\begin{equation}
\label{eq:kernel}
    \pi_x(z) = \frac{(m - 1)}{{m \choose \left|z\right|} \left|z\right| (m - \left|z\right|)},
\end{equation}
\begin{equation}
\label{eq:loss}
    L(f, g, \pi_x) = \sum_{z \in \left\{0, 1\right\}^m} \left[f(h_x(z)) - g(z)\right]^2 \cdot \pi_x(z),
\end{equation}
where $\left|z\right|$ is the number of non-zero elements of $z$, and $L$ is the squared error loss used for learning $g$.

Although adopted by the ML community, KernelSHAP is not fit for sequential settings. KernelSHAP's perturbations only affect the explained instance, ignoring the model's hidden state, which causes a mismatch between the data KernelSHAP attributes importance to and the data the model actually relies on.

To adapt KernelSHAP to the sequential setting, we discuss three possible approaches. The first consists in freezing the hidden state and perturb only the current instance. The resulting attributions would only explain the contribution of each of the feature values of the current instance, disregarding the rest of the sequence. Ho et al.~\cite{ho2020interpreting} followed this approach to explain RNN predictions on an ICU mortality dataset~\cite{MIMIC}.
%

The second consists in treating the hidden state as a feature and perturb it in block together with the actual features of the current instance. It would result in attributions to each feature of the current instance plus a single attribution value to the whole previous sequence (through the hidden state attribution), providing no granularity between past events. Finally, the third (followed in this work) consists in defining sequence-wide perturbations, enabling the calculation of attributions for features based on a full sequence of events (instances), as well as, event-level and cell-level (event-feature) attributions.


\section{TimeSHAP}
\label{sec:Method}
The main goal of our work is to develop an explainer that is both model-agnostic, post-hoc, and tailored for the recurrent setting. In order to explain a prediction, our method must provide both event and feature attributions throughout the sequence. At the same time, this method should be resource-efficient to be applicable in real-world scenarios.
Additionally, we aim to explain sequential models while preserving the three desirable properties of importance attribution stemming from the Shapley values: \textit{local accuracy}, \textit{missingness}, and \textit{consistency}~\cite{Shapley1953value}.

%


Hence, we put forth TimeSHAP, a model-agnostic recurrent that builds upon KernelSHAP~\cite{SHAPNIPS2017_7062}, and extends it to work on sequential data.
TimeSHAP attributes an importance value to each feature/event in the input that reflects the degree to which that feature/event affected the final prediction. 
In order to explain a sequential input, $X \in \mathbb{R}^{d \times l}$, with $l$ events and $d$ features per event, our method fits a linear explainer $g$ that approximates the local behavior of a complex explainer $f$ by minimizing the loss given by Equation~\ref{eq:loss}.
As events are simply features in a temporal dimension, and the algorithm for explaining features $x\in\mathbb{R}^{1 \times l}$ and events $x\in\mathbb{R}^{d \times 1}$ is conceptually equal, we will henceforth use the word \textit{feature} to mean both rows and columns of $X \in \mathbb{R}^{d \times l}$.
Thus, the formula for $g$ is

\begin{equation}
\label{eq:linear_model}
    f(h_X(z)) \approx g(z) = w_0 + \sum_{i=1}^{m}w_i z_i, 
\end{equation}
where the bias term $w_0 = f(h_X(\mathbf{0}))$ corresponds to the model's output with all features toggled off (dubbed \textit{base score}), the weights $w_i, i \in \left\{1, \dots, m\right\}$, correspond to the importance of each feature, and either $m=d$ or $m=l$, depending on which dimension is being explained.
The perturbation function $h_X: \{0, 1\}^m \mapsto \mathbb{R}^{d \times l}$ maps a coalition $z \in \left\{0, 1\right\}^{m}$ to the original input space $\mathbb{R}^{d \times l}$.
Note that the sum of all feature importances corresponds to the difference between the model's score $f(X) = f(h_X(\mathbf{1}))$ and the base score $f(h_X(\mathbf{0}))$.

\subsection{Sequence Perturbations}
\label{sec:TimeSHAP}

Input perturbations are generated depending on which dimension is being explained.
The perturbation function described in Equation~\ref{eq:perturbation} is suited to explain a single dimension of features.
We extend this function to the recurrent (and bi-dimensional) setting as follows.
Given a matrix $B \in \mathbb{R}^{d \times l}$ representing an uninformative input (the absence of discriminative features or events), 
a perturbation $h^{f}_X$ along the features axis (the rows) of the input matrix $X \in \mathbb{R}^{d \times l}$ is the result of mapping a coalition vector $z \in \{0, 1\}^d$ to the original input space $R^{d \times l}$, such that $z_i=1$ means that row $i$ takes its original value $X_{i,:}$, and $z_i=0$ means that row $i$ takes the background uninformative value $B_{i,:}$ .

In our setting, we define the background matrix $B \in \mathbb{R}^{l \times d}$ as containing the average feature values in the training dataset,

\begin{equation}
\label{eq:background_matrix}
B = \begin{bmatrix}
    \overline{x_1} & \dots & \overline{x_1} \\
    \overline{x_2} & \dots & \overline{x_2} \\
    \vdots & \ddots & \vdots \\
    \overline{x_l} & \dots & \overline{x_l} \\
    \end{bmatrix}.
\end{equation}

Thus, when $z_i=0$ the feature $i$ is essentially toggled off for all events of the sequence.
This is formalized by,

\begin{equation}
\label{eq:feature_perturbations}
h^{f}_X(z) = D_z X + (I - D_z) B, \quad D_z = \mathrm{diag}(z). \\
\end{equation}

On the other hand, a perturbation $h^{e}_X$ along the events axis (the columns) of the input matrix $X \in \mathbb{R}^{d \times l}$ is the result of mapping a coalition vector $z \in \{0, 1\}^l$ to the original input space $R^{d \times l}$, such that $z_j=1$ means that column $j$ takes its original value $X_{:,j}$, and $z_j=0$ means that column $j$ takes the value $B_{:,j}$ .
Thus, when $z_j=0$ all features of event $j$ are toggled off.
This is formalized by,

\begin{equation}
\label{eq:event_perturbations}
h^{e}_X(z) = X D_z + B (I - D_z), \quad D_z = \mathrm{diag}(z). \\
\end{equation}

Hence, when explaining features, $h_X = h^{f}_X$, whereas when explaining events, $h_X = h^{e}_X$.
This change in the perturbation function is the sole implementation difference between explaining events and features.


\begin{algorithm*}[t]
\caption{Temporal Coalition Pruning}
\label{algorithm:coalition_pruning}
\begin{flushleft}
\KwInput~input sequence $X$,  model to explain $f$,  tolerance $\eta$,
\end{flushleft}

\begin{algorithmic}[1]

\For{$i \in \{l - 1, l - 2, \dots, 1\}$} \Comment{Starting from the end of the sequence}
    \State $Z \gets \{[0, 0], [0, 1], [1, 0], [1, 1]\}$     \Comment{Full set of coalitions to use for each $i$}


    \State $w_1, w_2 \gets$ \textit{KernelSHAP}(            \Comment{Call adapted KernelSHAP}
        
        \qquad \textit{model}=$f$,

        \qquad \textit{input}=$[X_{:, 1:i}, X_{:, i+1:l}]$,    \Comment{$X$ composed of only two features}

        \qquad \textit{perturbation}=$h^{e}_X$,   \Comment{Parameterized by our temporal perturbation function}

        \qquad \textit{coalitions}=$Z$)         \Comment{Employing only $2^2$ coalitions (SHAP sees only 2 features)}

    \If{ $\left|w_1\right| < \eta$ }        \Comment{$w_1$ is the aggregate importance of all events up to $i$}
        \State \Return i    \Comment{Index from which it is safe to lump event importances}
    \EndIf
\EndFor
\State \Return 0    \Comment{No sequential group of events fits the pruning criteria}
\end{algorithmic}
\end{algorithm*}

Moreover, the perturbation of $X$ according to a null-vector coalition $z=\mathbf{0}$ is the same regardless of which dimension is being perturbed, $h^{f}_X(\mathbf{0}) = h^{e}_X(\mathbf{0})$, and equally for $z=1$, $h^{f}_X(\mathbf{1}) = h^{e}_X(\mathbf{1})$.

\subsection{Pruning}
\label{subsec:pruning}
One glaring issue with TimeSHAP is that the number of event (temporal) coalitions scales exponentially with the length of the observed sequence, just as in KernelSHAP the number of feature coalitions scales exponentially with the number of input features.
Moreover, in a recurrent setting, the input sequence can be arbitrarily long. We address this issue by proposing a temporal coalition pruning algorithm.

In a real-world scenario, it is common for events to be preceded by a long history of past events, with only a few of them being relevant to the current prediction (e.g., the whole transaction history of a client to detect fraud on the most recent one). Additionally, recurrent models are known to seldom encode information from events in the distant past~\cite{RNNvanishinggradient}.

With the previously stated insight, we group together older (unimportant) events as a single coalition of events, thereby reducing the number of coalitions by a factor of $2^{i - 1}$, where $i$ is the number of grouped events.
Essentially, we sacrifice explanation granularity on older, unimportant events, in favor of an exponential runtime speedup and increased precision of explanations on important events.
By grouping these events as one, we still enable all events to be considered in the explanation, but the importance of grouped events will be summed into a single Shapley value.

%

The pruning method, defined in Algorithm~\ref{algorithm:coalition_pruning}, consists in splitting the input sequence $X \in \mathbb{R}^{d \times l}$ into two sub-sequences $X_{:, 1:i}$, $X_{:, i+1:l}$, $i \in \{1, \dots, l - 1\}$ ($X_{:,l}$ being the most recent event), and computing the true Shapley values for each.
Computing these Shapley values amounts to considering $2^2=4$ coalitions.
We aim to find the largest $i$ such that the importance value for $X_{:, 1:i}$ falls below a given importance threshold $\eta$.

The computational cost of this pruning algorithm scales only linearly with the number of events, $\mathcal{O}({l})$.
Consequently, when employing pruning, the run-time of TimeSHAP is reduced from $\mathcal{O}({2^l})$ to $\mathcal{O}({2^{l-i}})$ with $i$ being the number of events lumped together. As older events are usually unimportant, a high number of events is pruned, thus $l-i \ll l$, rendering the performance increase highly significant.
Taking the best-case scenario of a model whose run-time scales linearly with $l$, such as a recurrent neural model (e.g., RNN, LSTM, GRU), TimeSHAP's run-time with temporal-coalition pruning (Algorithm~\ref{algorithm:coalition_pruning}) is $\mathcal{O}({l \cdot 2^{l-i}})$.

\subsection{Cell-Level Explanations}
\label{subsec:cell_level}

In addition to feature- and event-wise explanations, TimeSHAP also computes cell-wise attributions. These follow the same rationale as feature- and event-level explanations, perturbing cells of the input matrix to obtain their attribution.
As the number of cells is the product of the number of events by the number of features, even a small input grid represents an intractable task, $\mathcal{O}({2^{l d}})$, where $l$ and $d$ represents the number of events and features respectively. To obtain reliable cell-level explanations, either the number of considered cells needs to be drastically reduced or a pruning/grouping strategy needs to take place.

\begin{figure}[b]
	\centering
    \includegraphics[width=0.98\columnwidth]{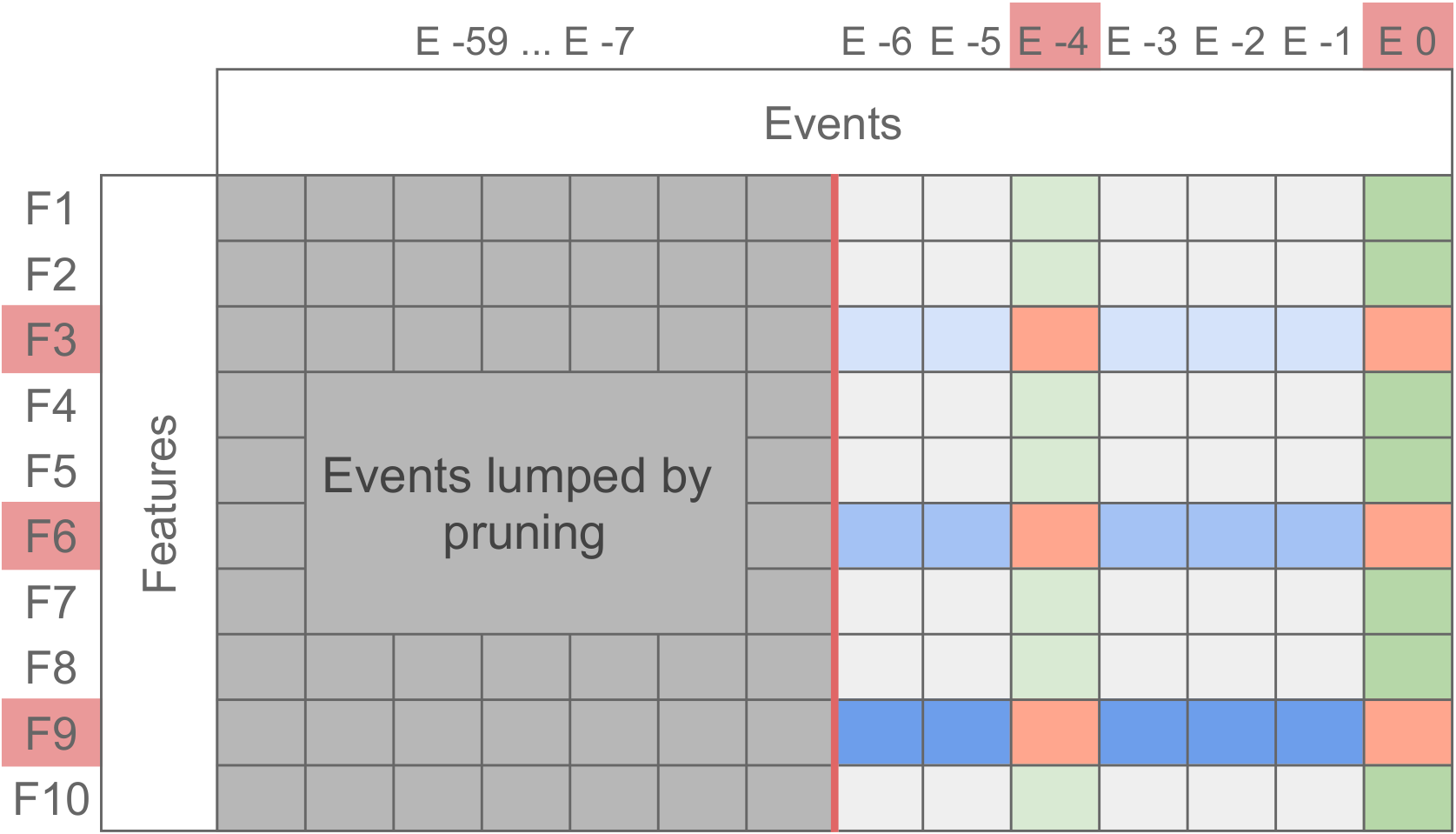}
	\caption{Cell-level grouping strategy, illustrated on a hypothetical input matrix with $60$ events and $10$ features. The pruning algorithm grouped the first $53$ events together. Events $-4$ and $0$, and features $3$, $6$, and $9$ were considered relevant (Shapley value greater than $\theta$).}
	\label{fig:cell_groups}
\end{figure}

We propose a cell grouping strategy to provide a spectrum of explanation granularity, while maintaining reliability of explanations and lowering computational complexity.
Similarly to Algorithm~\ref{algorithm:coalition_pruning}, cells are grouped together functioning as a single coalition.
A feature or event is considered relevant, if their absolute attribution is greater than a user defined threshold $\theta$.

To construct these groups, TimeSHAP first groups older events with coalition pruning.
Let $P$ be the set of cells of pruned events (marked with dark grey in Figure~\ref{fig:cell_groups}).
Afterwards, TimeSHAP computes 
both event- and feature-level explanations to find the most relevant columns (events) and rows (features).
We postulate that the most important cells of the input matrix are contained in the intersection of these most relevant rows and columns.

Given the set of most relevant features, $F^*$, and events, $E^*$, TimeSHAP isolates the cells at the intersection of these rows and columns as the most relevant cells, $C^*$ (marked with red in Figure~\ref{fig:cell_groups}):

\begin{equation}
    C^* = \left\{ ( f \cap e ) : (f, e) \in F^* \times E^* \right\} .
\end{equation}
Next, TimeSHAP forms groups of cells that belong to relevant events but not relevant features, $E^\prime$ (marked with shades of green in Figure~\ref{fig:cell_groups}):

\begin{equation}
    E^\prime = \left\{ ( e \cap \overline{F^*} ) : e \in E^* \right\} ,
\end{equation}
and groups of cells that belong to relevant features but not relevant events, $F^\prime$ (marked with shades of blue in Figure~\ref{fig:cell_groups}):

\begin{equation}
    F^\prime = \left\{ ( f \cap \overline{E^*} ) : f \in F^* \right\} .
\end{equation}
Finally, all remaining cells form a single coalition, $C^\prime$ (marked with light grey in Figure~\ref{fig:cell_groups}):
\begin{equation}
    C^\prime = \Omega \cap \overline{(P \cup C^* \cup E^\prime \cup F^\prime)} ,
\end{equation}
where $\Omega$ is the set of all cells.

The individual coalitions to be perturbed are the groups defined by $U = P \cup C^* \cup E^\prime \cup F^\prime \cup C^\prime$.
Consequently, TimeSHAP perturbs $k = |F^*| \cdot |E^*| + |F^*| + |E^*| + 2$ cell groups, a drastic reduction in the number of coalitions.


Let $m : z \mapsto U$ be a one-to-one mapping of elements of the coalition vector $z \in \{0, 1\}^{k}$ to elements of $U$, 
and $Z \in \{0, 1\}^{l \times d}$ be the coalition matrix defined by $Z_{r,c} = z_i \; \forall (r, c) \in m[z_i] \; \forall z_i \in z$.
The perturbation function $h_X^c$ is given by

\begin{equation}
    h_X^c(Z) = X \odot Z + B \odot (J - Z), \\
\end{equation}
where $J \in \{1\}^{l \times d}$, and $B \in \mathbb{R}^{d \times l}$ represents an uninformative input (the absence of discriminative features or events).

%
%



Through the aforementioned cell-grouping strategy, TimeSHAP computes explanations on a spectrum of granularity, isolating the individual attribution of cells deemed relevant, $C^*$, the attribution for the cells belonging to each relevant event/feature excluding intersection cells, $E^\prime$ and $F^\prime$, as well as the attribution for pruned cells and all other cells, which are believed to be less relevant.

\section{Case Study} 
\label{sec:case_study}

\begin{table*}[t]
\centering
\caption{Temporal-coalition pruning analysis (Algorithm~\ref{algorithm:coalition_pruning}). Sequence length indicates the number of events to be explained (events that were aggregated after pruning count as 1).}
\label{table:coalition_pruning_ato}
\begin{tabular}{lccccccc}
\toprule
&\textit{Original}& \text{$\eta=.005$} & \text{$\eta=.0075$} & \text{$\eta=.01$} & \text{$\eta=.025$} & \text{$\eta=.05$} \\
\midrule
Average seq. length & 182.1  &	69.0& 58.4 & 50.0 & 32.9 & 19.7\\
Median seq. length &  138.5  &  33.0 & 27.0   & 23.0  & 14.0  & 9.0   \\
Max seq. length  & 2187 &  2171 & 1376   & 1132 & 1130 & 879	 \\
Percentile at $log_2(32000)$ & 10.0 & 27.3 & 32.7 & 36.5 & 58.3 & 78.8 \\
TimeSHAP RSD, $\frac{\sigma}{\mu}$ & 1.71 & 1.19 & 1.17 & 1.09 & 0.98 & 0.68 \\
\bottomrule
\end{tabular}
\end{table*}

To validate our method, we use it to explain predictions of a GRU-based model on a real-world account takeover fraud detection task. Account takeover fraud is a form of identity theft where a fraudster gains access to a victim's bank account, enabling them to place unauthorized transactions~\cite{FraudSurvey2016,fbi_fraud2016}. The model used is composed of an embedding layer for categorical variables, followed by a GRU layer~\cite{Cho2014}, followed by a feed-forward classifier. This model obtained recall of $84.3\%$ at $1\%$ false positive rate (FPR) on the validation set, and a recall of $79.9\%$ at $0.89\%$ FPR on the test set.

The data is tabular, consisting of approximately 20M instances. Each instance, dubbed from here on as an event, represents one of three behaviours, or event types: transaction, where a client performs a monetary transaction; login, representing a client login on the banking application or website; or enrollment, representing account settings behaviours, such as logging into a new device or changing the password. Together with the information related to the type of event, each instance has the corresponding geo-location and demographics data associated. 
Due to the nature of each of these events, they have highly unbalanced statistics. Logins compose around $70\%$ of the dataset, $20\%$ are transactions and, enrollments make the last $10\%$.

We use TimeSHAP to explain all sequences that contain a positive prediction.
We set the maximum number of coalition samples to $\textit{n\_samples}=32K$, as it represents the best trade-off between TimeSHAP's computational cost and reliability. A higher number of sampled coalitions represents a heavy increase on the computational cost without rendering significant benefits on explanation variance. The same rationale was applied to choose the cell-level tolerance threshold of $\theta=0.1$, as this value assured that the most significant events and features were considered on the explanations while maintaining the number of possible coalitions low.
Additionally, we employ our proposed temporal-coalition pruning algorithm with tolerance $\eta=0.025$.

\subsection{Pruning Results} 
\label{subsec:chap5_pruning_ato}

For the pruning results, we randomly select $1000$ sequences with a positive prediction, and apply TimeSHAP together with Algorithm~\ref{algorithm:coalition_pruning}. Only event-level explanations were considered, as feature-level explanations are not affected by the pruning.

Table~\ref{table:coalition_pruning_ato} details average, median, and maximum number of events for unpruned sequences, and for sequences pruned with varying pruning tolerances, $\eta \in \{0.005, 0.0075, 0.01, 0.025, 0.05\}$.
The percentage of sequences whose length, $\left|X\right|$, is under $log_2(\textit{n\_samples}) \approx 15$ is shown in the fourth row. This represents the percentage of input sequences whose Shapley values can be exactly computed by exhaustively evaluating all $2^{\left|X\right|}$ coalitions. The Shapley values for all sequences longer than $log_2(\textit{n\_samples})$ are estimated by randomly sampling coalitions.

We note that, for the original sequences, exact Shapley values can only be computed for $10\%$ of the samples.
For the median original sequence, with a total number of coalitions on the order of $2^{\left|X\right|}=2^{139}$, 32K sampled coalitions represents $10^{-36}\%$ of the total universe of coalitions.
Hence, pruning is not only resource-efficient but also a necessary step in order to achieve accurate results.

When using $\eta=0.01$, we can compute exact Shapley values for $36.5\%$ of the input samples; while for $\eta=0.025$, we can compute exact values for $58.3\%$ of the input samples.
Hence, we choose $\eta=0.025$ as our pruning tolerance, providing a balance between explanations' consistency, run-time, and granularity.
The last row of Table~\ref{table:coalition_pruning_ato} shows the relative standard deviation\footnote{A standardized measure of dispersion, computed as the ratio of the standard deviation to the mean, $\frac{\sigma}{\mu}$.}~\cite{Brown1998} (RSD) of the Shapley values obtained over 10 runs of TimeSHAP for different pruning levels.
As expected, lower pruning tolerances (lower $\eta$) lead to finer-grained event-level explanations (higher number of explained events) but with higher variance (higher RSD values).
In fact, there is a strict negative relation between pruning tolerance and RSD values.
Running TimeSHAP on the original sequences leads to very high variance (RSD $1.71$), while the highest pruning tolerance $\eta=0.05$ leads to relatively low variance (RSD $0.68$).
Once again, $\eta=0.025$ (RSD $0.98$) achieves a well-balanced combination of metrics.

\subsection{Global Explanations}
\label{subsec:chap5_global_ato}



Supplying a data scientist with global explanations provides an overview of the model's decision process, revealing which features, or, in the case of TimeSHAP,  events, are relevant to the model and which ones are not.
TimeSHAP provides this type of explanations via aggregations and visualizations of local explanations.

We compute global explanations by applying TimeSHAP to all sequences that contained a positive prediction, and explaining the first positive prediction of each sequence (hence referred as $t=0$). We use $\eta=0.025$ as the tolerance of the temporal coalition pruning algorithm.

\begin{figure}[t]
	\centering
    \includegraphics[width=1.0\columnwidth]{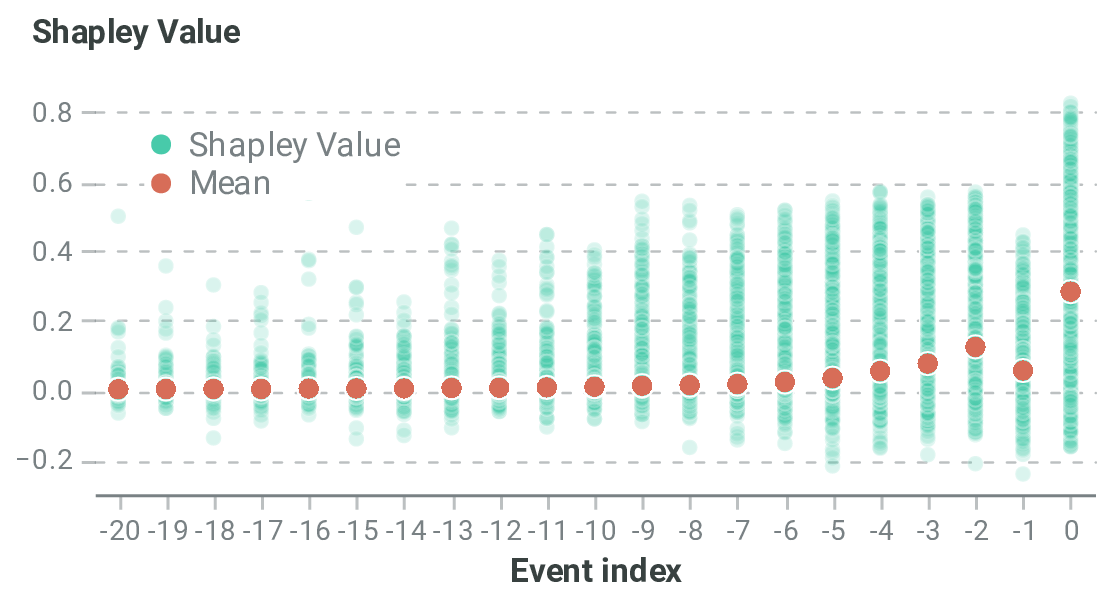}
	\caption{Event Shapley values density over the sequence index. The event 0 corresponds to the current input event. As the model only scores transactions, the preceding event is a login, having lower importance.
	}
	\label{fig:ato_global_event}
\end{figure}

Figure \ref{fig:ato_global_event} shows the global event-wise explanations.
From analyzing this plot we conclude that, on average, the latest transaction, $t=0$, is the event that most contributes to the positive prediction (average contribution of $0.28$).
The next most relevant events are those between indices $-4$ and  $-1$, with average contributions ranging from $0.03$ to $0.13$.
For events with an index lower than $-5$, the average contribution is around $0$. Nonetheless, it is possible to observe that there is still a significant number of sequences that have events between indices $-5$ and $-20$ with high contributions.
This long tail of distant events with significant contributions to the predicted score supports the model's ability of keeping crucial information in the hidden state and carrying it across time-steps, as well as TimeSHAP's ability to capture it.

We found that events at index $t=-1$ have a lower importance ($0.06$) than adjacent events, due to the fact that these events are often logins that precede the explained transaction and, most of the times, logins have a low attribution value given by TimeSHAP.
One crucial insight provided by these explanations is that the most recent event is responsible for, on average, $41$\% of the sequence’s score, while the preceding events represent $59$\% of the sequence’s score. This shows TimeSHAP's ability to understand the sequential domain, as other post-hoc methods would attribute $100$\% to the most recent event.

\begin{figure}[t]
	\centering
    \includegraphics[width=1.0\columnwidth]{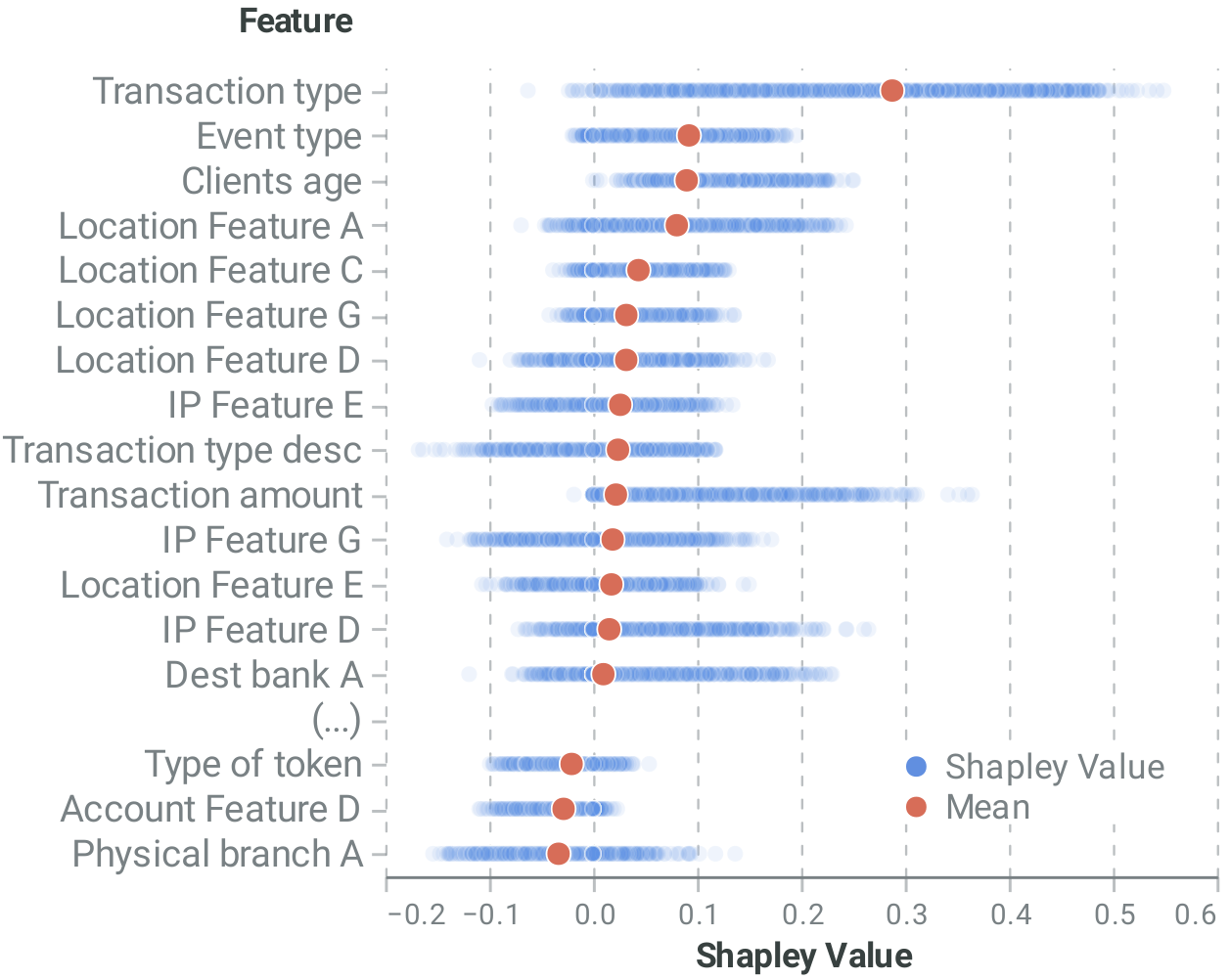}
	\caption{Feature Shapley values density over the model features. Only features with an average absolute contribution greater than $0.01$ are displayed. Feature names are obfuscated due to security reasons.
	}
	\label{fig:ato_global_feat}
\end{figure}

Figure \ref{fig:ato_global_feat} shows a plot of the global feature-wise explanations.
We observe that some features have predominantly positive contributions, meaning that the model routinely relies on these features to predict account takeover fraud.
These include the Transaction ($0.29$) and Event ($0.092$) types, the clients' age ($0.090$), and features related to the IP and location of the events ($0.08$ to $0.03$). 

Event type indicates if the event is a transaction, login, or enrollment; while Transaction type is a feature used solely on transaction events to indicate more specific transaction taxonomy (e.g., transaction to a different bank or transaction between accounts of the same client).
These features, indicated as the most relevant ones by TimeSHAP, agree with domain knowledge where the event and transaction types together with IP and location features encode account behavior, while the client's age correlates with their susceptibility to fraud. We later conducted a bias audit and observed a disparity on false positive rates for older clients.
%

On the other hand, some features have an average negative contribution to the score. These features are related to the account (physical branch A and account feature D) and the type of token of the event in question.
These features are related to the authentication and security of the account and the event itself. 
This negative contribution is in accordance with our domain knowledge, as these feature represent the authentication and security of the client and, therefore, provide confidence in the legitimacy of the event.


Additionally, feature-wise global explanations show two features with a null contribution for all observed sequences. This null contribution might stem from our choice of a background/uninformative event, or it may indicate that the model is not using the feature as a predictor. Upon inspection of the raw data we see that these features take a single constant value for all sequences, demonstrating the usefulness of TimeSHAP in aiding data scientists to iterate and debug models. 


\subsection{Local explanations}
\label{sec:individual_exp}

\begin{figure*}
     \centering
     \begin{subfigure}[b]{0.33\textwidth}
         \centering
         \includegraphics[height=1.5in]{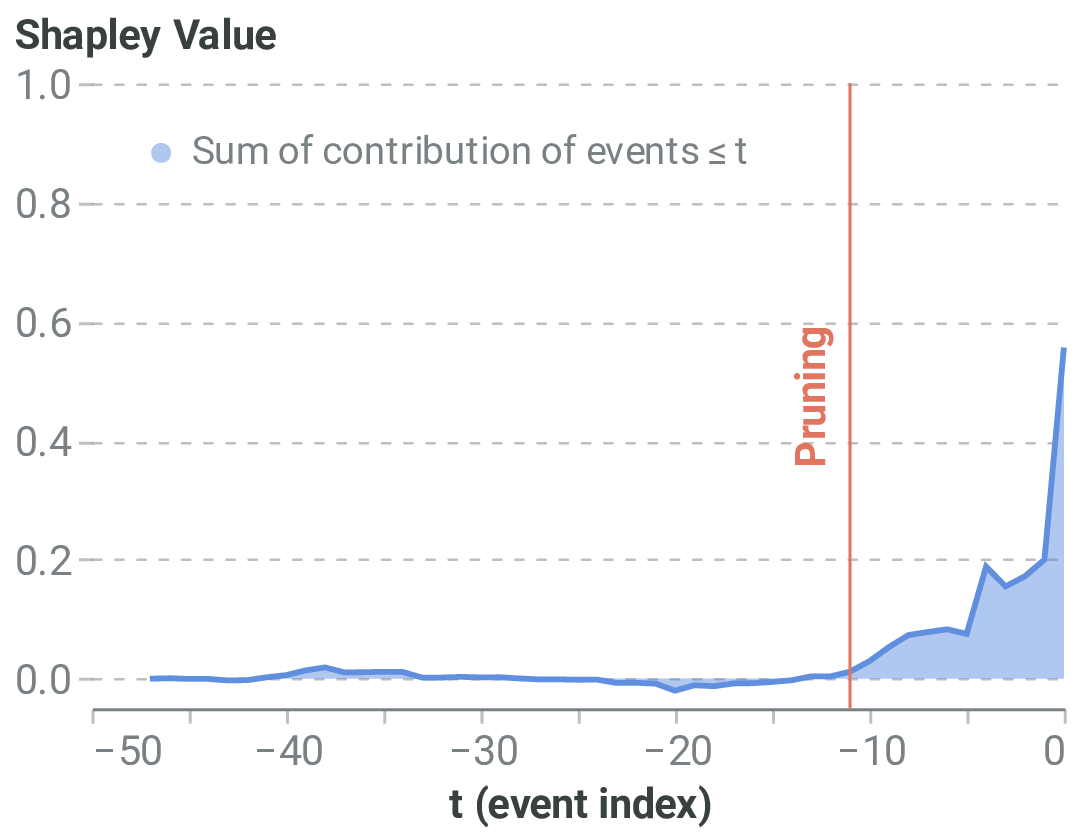}
         \caption{}
         \label{fig:h_vs_x_1}
     \end{subfigure}
     \hfill
     \begin{subfigure}[b]{0.33\textwidth}
         \centering
         \includegraphics[height=1.5in]{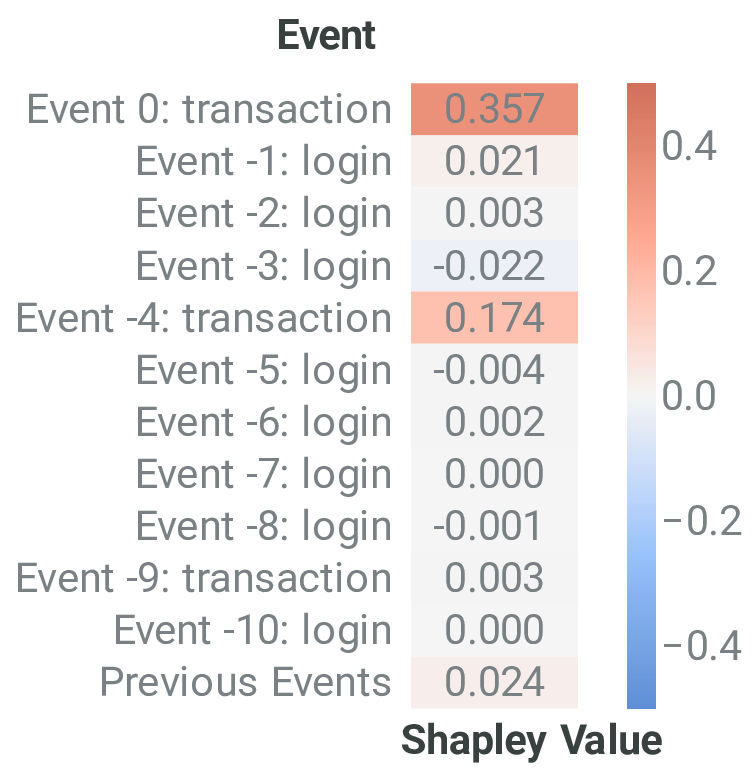}
         \caption{}
         \label{fig:event_1}
     \end{subfigure}
     \hfill
     \begin{subfigure}[b]{0.33\textwidth}
         \centering
         \includegraphics[height=1.5in]{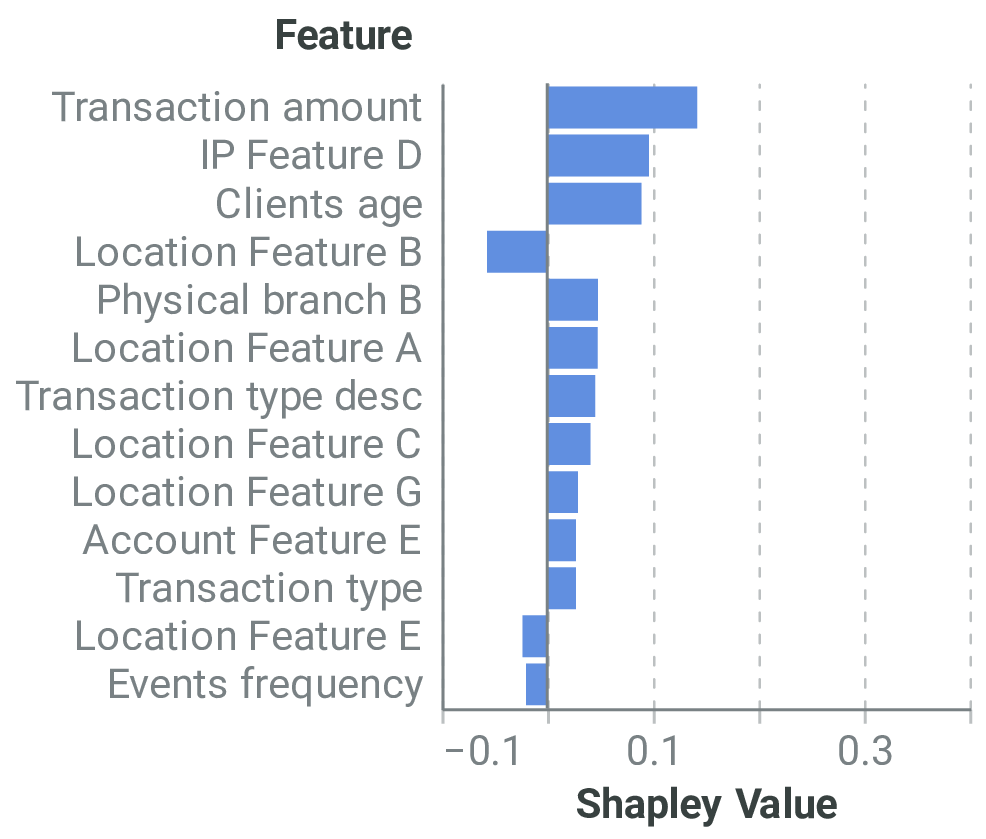}
         \caption{}
         \label{fig:feat_1}
     \end{subfigure}

     \hfill
     \begin{subfigure}[b]{0.33\textwidth}
         \centering
         \includegraphics[height=1.5in]{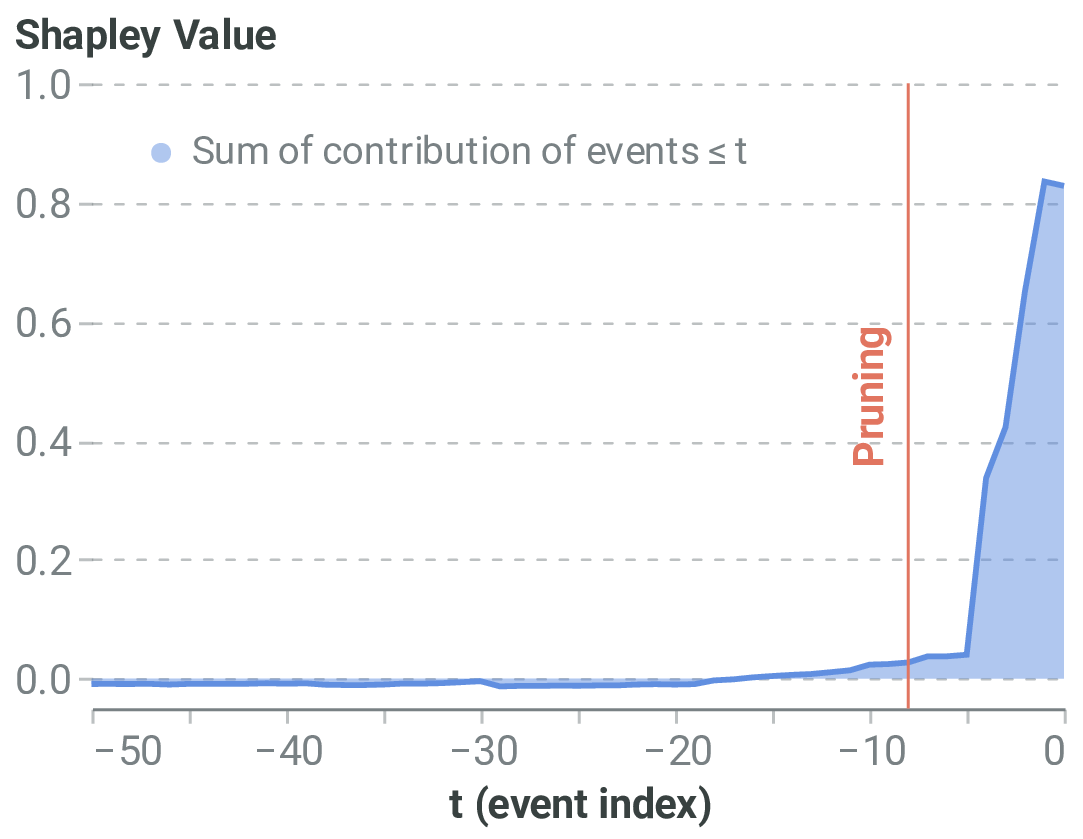}
         \caption{}
         \label{fig:h_vs_x_2}
     \end{subfigure}
     \hfill
     \begin{subfigure}[b]{0.33\textwidth}
         \centering
         \includegraphics[height=1.5in]{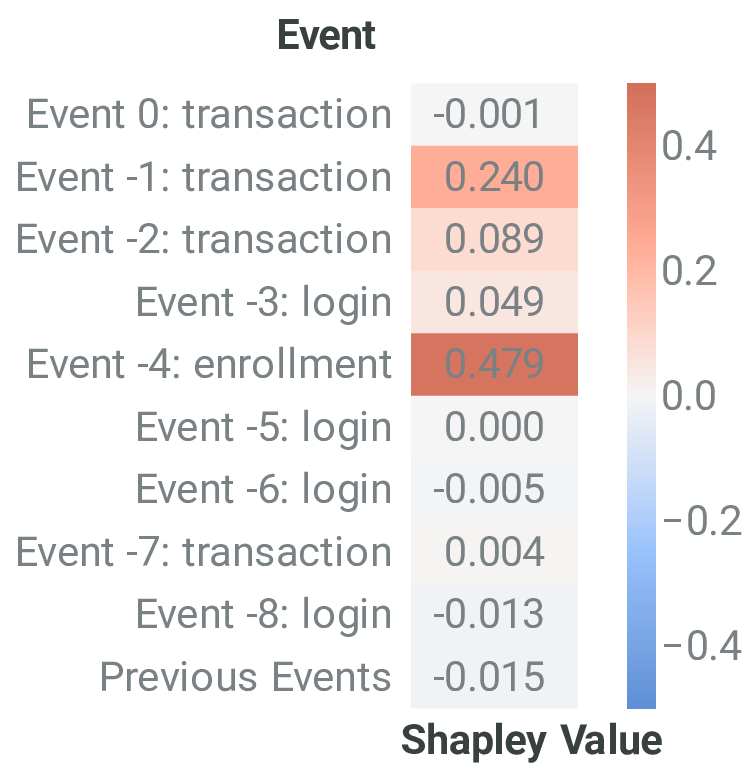}
         \caption{}
         \label{fig:event_2}
     \end{subfigure}
     \hfill
     \begin{subfigure}[b]{0.33\textwidth}
         \centering
         \includegraphics[height=1.5in]{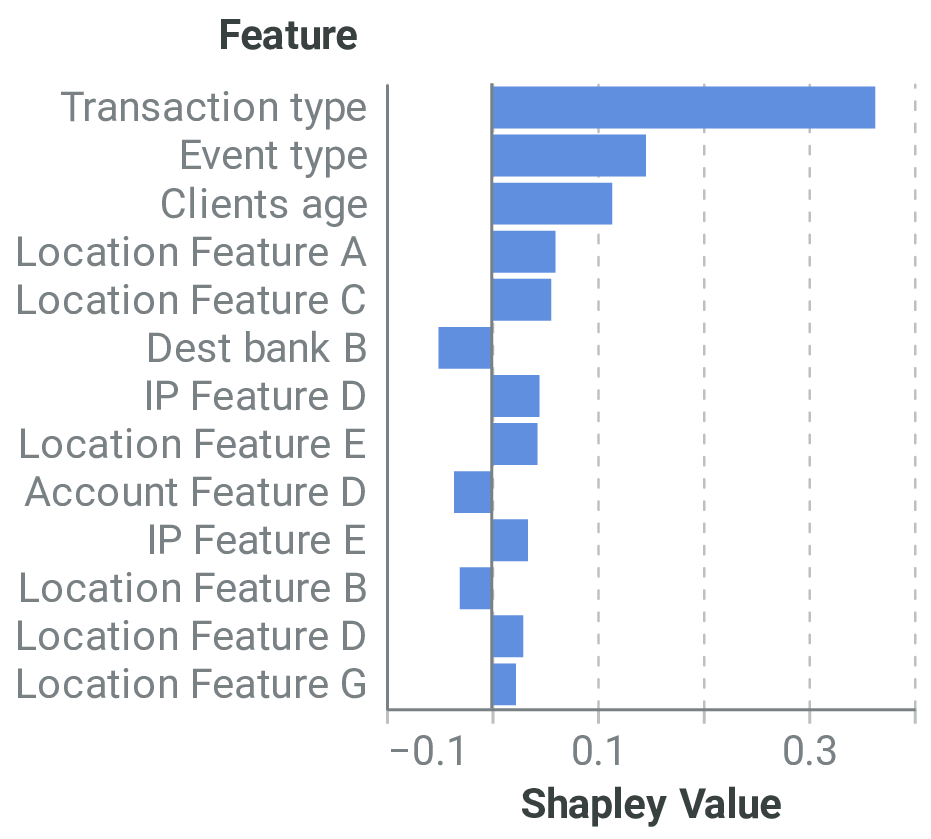}
         \caption{}
         \label{fig:feat_2}
     \end{subfigure}
     
     \begin{subfigure}[b]{0.48\textwidth}
         \centering
         \includegraphics[height=1.2in]{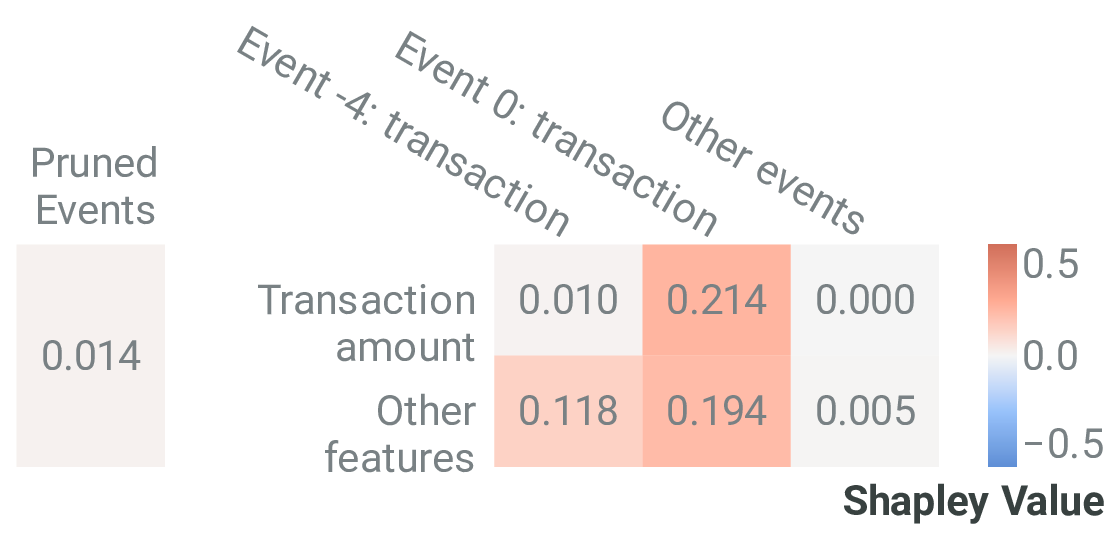}
         \caption{}
         \label{fig:cell_1}
     \end{subfigure}
     \hfill
     \begin{subfigure}[b]{0.48\textwidth}
         \centering
         \includegraphics[height=1.5in]{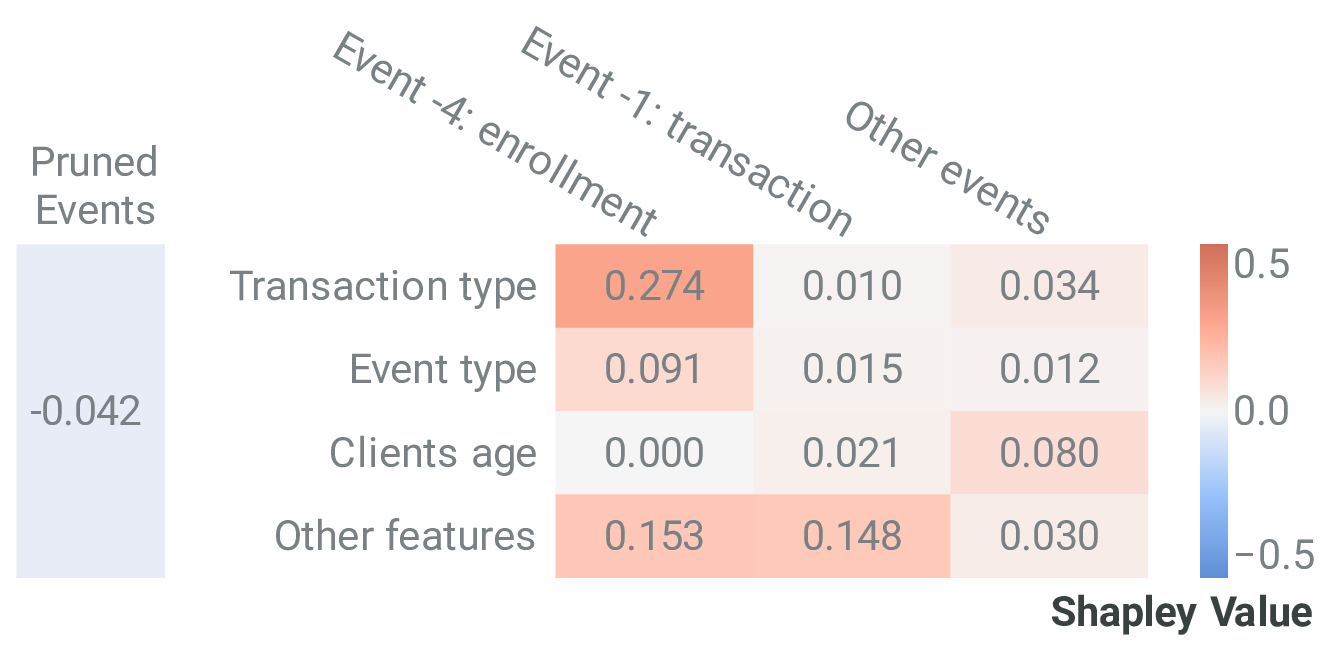}
         \caption{}
         \label{fig:cell_2}
     \end{subfigure}

    \caption{
        Figures (a), (b), (c), and (g) show TimeSHAP results for Sequence A.
        Figures (d), (e), (f), and (h) show TimeSHAP results for Sequence B.
        Figures (a) and (d) show the importance of grouped events, $X_{:, : t}$, as $t$ changes, computed for Algorithm~\ref{algorithm:coalition_pruning}.
        Event-level importance is shown in Figures (b) and (e).
        Feature-level importance is shown in Figures (c) and (f).
        Cell-level importance shown in Figures (g) and (h). For cell-level calculations, features and events are considered if importance greater than $0.1$. 
    }
    \label{fig:local_explanations_ato}

\end{figure*}

Local explanations explain the model's rationale regarding one specific instance. These explanations can be used in several use-cases, for example, for bias auditing or model debugging. However, these explanations can mostly be used by end-users, the fraud analysts, to aid their decision-making tasks.

In this Section we present two case-studies/sequences. We selected two predicted-positive sequences, hence labeled A and B. Sequence A has a model score of $f(A)=0.57$, and a total length of 47 events. Sequence B has a model score of $f(B)=0.84$, and a total length of 286 events.
As a convention, we dub the current event's index (the most recent) as $t=0$, and use negative indices for previous events (the event at index $t=-1$ immediately precedes the event at index $t=0$, and so on).

Figures~\ref{fig:h_vs_x_1}, \ref{fig:h_vs_x_2}, show the Shapley values, (importance,) respectively, for Sequences A and B split into two disjoint sub-sequences at a given index $t$.
This corresponds to the application of the ``\textit{for loop}'' in Algorithm~\ref{algorithm:coalition_pruning}, continuing even after the pruning condition has been fulfilled. Figure~\ref{fig:h_vs_x_2} only displays the first 50 indices as displaying all 286 events would clutter the Figure.
As expected, the aggregate importance of older events (from the beginning of the sequence up to index $t$) suffers a steep decrease as its distance to the current event increases.
This trend corroborates our hypothesis and supports our coalition pruning algorithm.
Using coalition pruning tolerance $\eta=0.025$, sequence A is pruned to 11 events, grouping the 36 older events together, and sequence B is pruned to 9 events, grouping the last 277 events together.


Event-wise explanations for sequence A are shown in Figure~\ref{fig:event_1}, and its feature-wise explanations in Figure~\ref{fig:feat_1}.
We conclude that there are two events crucial for the model's prediction: the transaction being explained ($t=0$), with a Shapley value of 0.36, and another transaction, 4 events before ($t=-4$), with a Shapley value of 0.17.
Between the two relevant transactions (events $-3 \leq t \leq -1$), there are three logins with little to no importance.
Prior to event $t=-4$, there are 5 logins and one transaction with reduced importance, which were nonetheless left unpruned by Algorithm~\ref{algorithm:coalition_pruning}.

Regarding feature importances, we observe that the most relevant features are, in decreasing order of importance, the transaction's \textit{amount}, \textit{IP feature D}, and the clients' \textit{age}.
When inspecting the raw feature data, we observe that the amount transferred at transaction $t=0$ is unusually high, a known account takeover indicator.
This is in accordance with the simultaneous high event importance for $t=0$, together with the high feature importance for the transaction amount.
Moreover, we observe that the client's age is relatively high, another well-known fraud indicator, as elderly clients are often more susceptible to being victims of fraud~\cite{10.2307/23859598}. When analyzing \textit{IP feature D}, although this feature does not show any strange behavior, it assumes a value that is frequent throughout the dataset. Upon further inspection, we conclude that the IP belongs to a cloud hosting provider, which domain experts confirm to be suspicious behavior.

Analyzing the cell-level explanations for this sequence, with $\theta=0.1$, we verify the previously stated insight, where the high transaction amount at $t=0$ is abnormal. This cell is solely responsible for $0.21$ of the score, revealing an extreme importance in a sequence with more than $2000$ cells. These cell-level explanations also show that the relevant cells are present in the relevant events, with the cells on all non-relevant events receiving a total importance of $0.005$.


Event-wise explanations for sequence B are shown in Figure~\ref{fig:event_2}, and its feature-wise explanations in Figure~\ref{fig:feat_2}.
Regarding event importance, we conclude that the most relevant events are at indices $-4$ and $-1$, with their corresponding Shapley values of $0.48$ and $0.24$, followed by events $-2$ ($0.089$) and $-3$ ($0.049$).
Interestingly, for this sequence, the most relevant event is not the current one ($t=0$), with near null contribution to the score ($0.001$).

The event types for the sequence of events from $t=-4$ to $t=-2$ are \textit{enrollment}-\textit{login}-\textit{transaction}, a well-known pattern that is repeated on numerous stolen accounts.
This sequence of events encodes a change of account settings, e.g., a password change (enrollment), followed by a login into the captured account, subsequently followed by one or more fraudulent transactions.
Interestingly, events $t=-1$ and $t=-2$ are transactions that succeed the fraudulent enrollment and login, but precede the current transaction ($t=0$).
The information up to $t=-1$ is already sufficient for the model to correctly identify the account as compromised, corroborated by the low contribution of the transaction at $t=0$.

Regarding feature importances, the most relevant features are related to the \textit{transaction type}, \textit{event type}, the clients' \textit{age}, and the \textit{location}.
%
When inspecting the raw feature data, we observe that the client is in the elderly age range, which, as previously mentioned, may indicate a more susceptible demographic. When analyzing the location features \textit{Location feature A} and \textit{Location feature D}, we observe a discrepancy between the location of the enrollment, login and transactions from the account's history. This discrepancy in physical location is highly suspicious and indicates that there was an enrollment on the account from a previously unused location.

Regarding the cell-level explanations, with $\theta=0.1$, we obtain that the most relevant cell, with importance $0.274$, is the intersection between the most relevant event and feature. This cell is followed by the groups of cells that aggregate the other features of the relevant events (separately) with the other features of event $t=-4$, having a relevance of $0.153$, and of event $t=-1$, with an importance of $0.148$. Considering the second most relevant feature, \textit{Event type}, it appears to be relevant at event $t=-4$ with importance $0.091$ and \textit{Customer age} being important on mostly other non-relevant events. Another relevant insight provided by these explanations is that the cells of non-relevant features or events have an importance of $0.03$. 

\section{Conclusion}
\label{sec:conclusion}

As in many sequential decision-making tasks, RNN-based models are state-of-the-art in fraud detection. However, in practice, financial institutions and regulators refrain from adopting these models due to concerns regarding their opaqueness. The ability to explain their predictions is essential to build trust among stakeholders, such as data scientists, fraud analysts or regulators. While considerable effort has been guided towards explaining deep learning models, recurrent models have received comparatively little attention.

In this work, we proposed TimeSHAP, a model-agnostic, post-hoc recurrent explainer.
Our method is suited to explain the predictions of any recurrent model, regardless of architecture, only requiring access to the features of each instance and an inference API.
TimeSHAP provides three types of explanations: event-, feature- and cell-level attributions, computed through perturbation functions tailored for sequences.
In addition, we proposed a pruning algorithm that both decreases the execution time of TimeSHAP and decreases the variance of the explanations. This way, data scientists can validate, debug and iterate model development, while humans-in-the-loop (e.g., fraud analysts) can better understand a model's prediction and decide accordingly to follow or to override it.

We used TimeSHAP to explain a real-world bank account takeover RNN-based model and to shed some light into its decision processes. We validate the model's reliability as both the features and events deemed important by the explanations were corroborated by domain experts, namely the enrollment-login-transaction fraud pattern that is typical in account takeover.
We find the event being explained is, on average, responsible for $41$\% of the sequence’s score, while the preceding events represent $59$\% of the sequence’s score. Moreover, we find that the client's age is an important feature for this model,
suggesting potentially discriminatory reasoning that was later confirmed in a bias audit.
\begin{acks}

The project CAMELOT (reference POCI-01-0247-FEDER-045915) leading to this work is co-financed by the ERDF - European Regional Development Fund through the Operational Program for Competitiveness and Internationalisation - COMPETE 2020, the North Portugal Regional Operational Program - NORTE 2020 and by the Portuguese Foundation for Science and Technology - FCT under the CMU Portugal international partnership.
MF was partially supported by Funda\c{c}\~ao para a Ci\^encia e Tecnologia, grant UIDB/50008/2020.
We would like to express our gratitude to Beatriz Malveiro and Jo\~ao Palmeiro for invaluable feedback on the paper's visualizations.
\end{acks}

\balance
\bibliographystyle{unsrt}
\bibliography{refs}

\begin{thebibliography}{10}

\bibitem{Simonyan2014DeepIC}
Karen Simonyan, Andrea Vedaldi, and Andrew Zisserman.
\newblock Deep inside convolutional networks: Visualising image classification
  models and saliency maps.
\newblock {\em CoRR}, abs/1312.6034, 2014.

\bibitem{denil2014extraction}
Misha Denil, Alban Demiraj, and Nando De~Freitas.
\newblock Extraction of salient sentences from labelled documents.
\newblock {\em arXiv preprint arXiv:1412.6815}, 2014.

\bibitem{li2016understanding}
Jiwei Li, Will Monroe, and Dan Jurafsky.
\newblock Understanding neural networks through representation erasure.
\newblock {\em arXiv preprint arXiv:1612.08220}, 2016.

\bibitem{MONTAVON2017211}
Grégoire Montavon, Sebastian Lapuschkin, Alexander Binder, Wojciech Samek, and
  Klaus-Robert Müller.
\newblock Explaining nonlinear classification decisions with deep {T}aylor
  decomposition.
\newblock {\em Pattern Recognition}, 65:211 -- 222, 2017.

\bibitem{10.1371/journal.pone.0130140}
Sebastian Bach, Alexander Binder, Grégoire Montavon, Frederick Klauschen,
  Klaus-Robert Müller, and Wojciech Samek.
\newblock On pixel-wise explanations for non-linear classifier decisions by
  layer-wise relevance propagation.
\newblock {\em PLOS ONE}, 10(7):1--46, 07 2015.

\bibitem{shrikumar2019learning}
Avanti Shrikumar, Peyton Greenside, and Anshul Kundaje.
\newblock Learning important features through propagating activation
  differences.
\newblock In Doina Precup and Yee~Whye Teh, editors, {\em Proc. of the 34th
  Int. Conf. on Machine Learning}, volume~70 of {\em Proc. of Machine Learning
  Research}, pages 3145--3153, Sydney, Australia, 2017.

\bibitem{lstmvis}
Hendrik Strobelt, Sebastian Gehrmann, Hanspeter Pfister, and Alexander~M. Rush.
\newblock {LSTMV}is: A tool for visual analysis of hidden state dynamics in
  recurrent neural networks.
\newblock {\em IEEE Transactions on Visualization and Computer Graphics},
  24(1):667--676, 2018.

\bibitem{murdoch2017automatic}
W.~James Murdoch and Arthur Szlam.
\newblock Automatic rule extraction from long short term memory networks.
\newblock {\em 5th Int. Conf. on Learning Representations, ICLR 2017}, 2017.

\bibitem{karpathy2015visualizing}
Matthew~D. Zeiler and Rob Fergus.
\newblock Visualizing and understanding convolutional networks.
\newblock In David Fleet, Tomas Pajdla, Bernt Schiele, and Tinne Tuytelaars,
  editors, {\em European Conf. on Computer Vision -- (ECCV) 2014}, pages
  818--833, 2014.

\bibitem{SHAPNIPS2017_7062}
Scott~M. Lundberg and Su-In Lee.
\newblock A unified approach to interpreting model predictions.
\newblock In {\em Proc. of the 31st Int. Conf. on Neural Information Processing
  Systems}, NIPS'17, page 4768–4777, 2017.

\bibitem{rumelhart1986learning}
David~E Rumelhart, Geoffrey~E Hinton, and Ronald~J Williams.
\newblock Learning representations by back-propagating errors.
\newblock {\em Nature}, 323(6088):533--536, 1986.

\bibitem{teddler2020javelin}
Krista Tedder and John Buzzar.
\newblock Identity fraud study: Genesis of the identity fraud crisis.
\newblock {\em Javelin Strategy \& Report}, 2020.

\bibitem{Hochreiter1997}
Sepp Hochreiter and J{\"{u}}rgen Schmidhuber.
\newblock {Long short-term memory}.
\newblock {\em Neural Computation}, 9(8):1735--1780, 1997.

\bibitem{Cho2014}
Kyunghyun Cho, Bart van Merrienboer, Dzmitry Bahdanau, and Yoshua Bengio.
\newblock {On the properties of neural machine translation: encoder–decoder
  approaches}.
\newblock In {\em Proc. of the Eighth Workshop on Syntax, Semantics and
  Structure in Statistical Translation {(SSST)}}, pages 103--111, 2014.

\bibitem{Ribeiro2016}
Marco~Tulio Ribeiro, Sameer Singh, and Carlos Guestrin.
\newblock Model-agnostic interpretability of machine learning.
\newblock {\em ICML Workshop on Human Interpretability in Machine Learning (WHI
  2016)}, jun 2016.

\bibitem{NIPS2019_9264}
Aya~Abdelsalam Ismail, Mohamed Gunady, Luiz Pessoa, Hector Corrada~Bravo, and
  Soheil Feizi.
\newblock Input-cell attention reduces vanishing saliency of recurrent neural
  networks.
\newblock In H.~Wallach, H.~Larochelle, A.~Beygelzimer, F.~d\textquotesingle
  Alch\'{e}-Buc, E.~Fox, and R.~Garnett, editors, {\em Advances in Neural
  Information Processing Systems 32}, pages 10814--10824. 2019.

\bibitem{Zhang_2018}
Jinghe Zhang, Kamran Kowsari, James~H. Harrison, Jennifer~M. Lobo, and Laura~E.
  Barnes.
\newblock Patient2vec: A personalized interpretable deep representation of the
  longitudinal electronic health record.
\newblock {\em IEEE Access}, 6:65333–65346, 2018.

\bibitem{RETAIN}
Edward Choi, Mohammad~Taha Bahadori, Jimeng Sun, Joshua Kulas, Andy Schuetz,
  and Walter Stewart.
\newblock Retain: An interpretable predictive model for healthcare using
  reverse time attention mechanism.
\newblock In D.~D. Lee, M.~Sugiyama, U.~Von. Luxburg, I.~Guyon, and R.~Garnett,
  editors, {\em Advances in Neural Information Processing Systems 29}, pages
  3504--3512. 2016.

\bibitem{attentionRNNs}
Ying Sha and May~D. Wang.
\newblock Interpretable predictions of clinical outcomes with an
  attention-based recurrent neural network.
\newblock In {\em Proc. of the 8th ACM Int. Conf. on Bioinformatics,
  Computational Biology, and Health Informatics}, ACM-BCB '17, page 233–240,
  New York, NY, USA, 2017.

\bibitem{serrano-smith-2019-attention}
Sofia Serrano and Noah~A. Smith.
\newblock Is attention interpretable?
\newblock In {\em Proc. of the 57th Annual Meeting of the Association for
  Computational Linguistics}, pages 2931--2951, July 2019.

\bibitem{attention-is-not-explanation}
Sarthak Jain and Byron~C. Wallace.
\newblock {A}ttention is not explanation.
\newblock In {\em Proc. of the 2019 Conf. of the North {A}merican Chapter of
  the Association for Computational Linguistics: Human Language Technologies,
  Volume 1 (Long and Short Papers)}, pages 3543--3556, June 2019.

\bibitem{attention-is-not-not-explanation}
Sarah Wiegreffe and Yuval Pinter.
\newblock Attention is not not explanation.
\newblock In {\em Proc. of the 2019 Conf. on Empirical Methods in Natural
  Language Processing and the 9th Int. Joint Conf. on Natural Language
  Processing (EMNLP-IJCNLP)}, pages 11--20, November 2019.

\bibitem{axiomatic}
Mukund Sundararajan, Ankur Taly, and Qiqi Yan.
\newblock Axiomatic attribution for deep networks.
\newblock In {\em Proc. of the 34th Int. Conf. on Machine Learning}, (ICML),
  page 3319–3328, 2017.

\bibitem{hochreiter1991untersuchungen}
Sepp Hochreiter.
\newblock Untersuchungen zu dynamischen neuronalen netzen.
\newblock {\em Diploma, Technische Universit{\"a}t M{\"u}nchen}, 91(1), 1991.

\bibitem{bbexplainers}
Riccardo Guidotti, Anna Monreale, Salvatore Ruggieri, Franco Turini, Fosca
  Giannotti, and Dino Pedreschi.
\newblock A survey of methods for explaining black box models.
\newblock {\em ACM Comput. Surv.}, 51(5), August 2018.

\bibitem{LIME}
Marco~Tulio Ribeiro, Sameer Singh, and Carlos Guestrin.
\newblock "{W}hy should {I} trust you?": Explaining the predictions of any
  classifier.
\newblock In {\em Proc of the 22nd ACM SIGKDD Int. Conf. on Knowledge Discovery
  and Data Mining}, (KDD), page 1135–1144, 2016.

\bibitem{Young1985}
H.~Peyton Young.
\newblock {Monotonic solutions of cooperative games}.
\newblock {\em Int. Journal of Game Theory}, 14(2):65--72, jun 1985.

\bibitem{Strumbelj2014}
Erik {\v{S}}trumbelj and Igor Kononenko.
\newblock {Explaining prediction models and individual predictions with feature
  contributions}.
\newblock {\em Knowledge and Information Systems}, 41(3):647--665, 2014.

\bibitem{ho2020interpreting}
Long~V. Ho, Melissa Aczon, David Ledbetter, and Randall Wetzel.
\newblock Interpreting a recurrent neural network’s predictions of {ICU}
  mortality risk.
\newblock {\em Journal of Biomedical Informatics}, 114:103672, 2021.

\bibitem{MIMIC}
Alistair~EW Johnson, Tom~J Pollard, Lu~Shen, H~Lehman Li-Wei, Mengling Feng,
  Mohammad Ghassemi, Benjamin Moody, Peter Szolovits, Leo~Anthony Celi, and
  Roger~G Mark.
\newblock {MIMIC-III}, a freely accessible critical care database.
\newblock {\em Scientific data}, 3(1):1--9, 2016.

\bibitem{Shapley1953value}
Lloyd~S Shapley.
\newblock A value for n-person games.
\newblock {\em Contributions to the Theory of Games}, 2(28):307--317, 1953.

\bibitem{RNNvanishinggradient}
Yoshua Bengio, Paolo Frasconi, and Patrice Simard.
\newblock The problem of learning long-term dependencies in recurrent networks.
\newblock In {\em IEEE Int. Conf. on Neural Networks}, volume~3, pages
  1183--1188, 1993.

\bibitem{FraudSurvey2016}
Aisha Abdallah, Mohd~Aizaini Maarof, and Anazida Zainal.
\newblock Fraud detection system: A survey.
\newblock {\em Journal of Network and Computer Applications}, 68:90 -- 113,
  2016.

\bibitem{fbi_fraud2016}
{Federal Bureau of Investigation}.
\newblock Credit card fraud.
\newblock
  \url{https://www.fbi.gov/scams-and-safety/common-scams-and-crimes/credit-card-fraud},
  Jun 2016.
\newblock Accessed: 2021-01-09.

\bibitem{Brown1998}
Charles~E. Brown.
\newblock {\em Coefficient of Variation}, pages 155--157.
\newblock 1998.

\bibitem{10.2307/23859598}
Jinkook Lee and Horacio Soberon-Ferrer.
\newblock Consumer vulnerability to fraud: Influencing factors.
\newblock {\em The Journal of Consumer Affairs}, 31(1):70--89, 1997.

\end{thebibliography}

\end{document}